\documentclass[letterpaper, 10pt, twocolumn]{article}
\usepackage{clean}
\shorttitle{Sampling- and Gradient-based Planning for Contact}

\usepackage{graphicx}
\usepackage{subfig}
\graphicspath{{figs/}}
\usepackage{amsmath,amssymb,amsfonts}
\usepackage{algorithm}
\usepackage{algorithmicx}
\usepackage{algpseudocode}
\usepackage{setspace}

\usepackage{xcolor}
\usepackage[colorlinks=true, linkcolor=black, urlcolor=cyan, filecolor=black, citecolor=black]{hyperref}
\usepackage{url}
\usepackage{amsmath,bm}
\usepackage{multicol, multirow, makecell} %
\usepackage{comment}

\usepackage{cite}
\title{Combining Sampling- and Gradient-based Planning for Contact-rich Manipulation}
\author{Filippo Rozzi$^1$,  Loris Roveda$^2$, and Kevin Haninger$^3$ %
\thanks{$^1$Corresponding author, Politecnico di Milano, Department of Mechanical Engineering, Milano, Italy  {\tt filippo.rozzi@mail.polimi.it}}
\thanks{$^2$Istituto Dalle Molle di Studi sull'Intelligenza Artificiale (IDSIA), Scuola Universitaria Professionale della Svizzera Italiana (SUPSI), Università della Svizzera Italiana (USI) IDSIA-SUPSI, Lugano, Switzerland {\tt loris.roveda@idsia.ch}}
\thanks{$^3$Department of Automation at Fraunhofer IPK, Berlin, Germany {\tt kevin.haninger@ipk.fraunhofer.de}}
\thanks{This project has received funding from the European Union's Horizon 2020 research and innovation programme under grant agreement 101058521 — CONVERGING.}}

\begin{document}
\maketitle
\begin{abstract}
Planning for contact-rich manipulation involves discontinuous dynamics, which presents challenges to planning methods. Sampling-based planners have higher sample complexity in high-dimensional problems and cannot efficiently handle state constraints such as force limits. Gradient-based solvers can suffer from local optima and their convergence rate is often worse on non-smooth problems. We propose a planning method that is both sampling- and gradient-based, using the Cross-entropy Method to initialize a gradient-based solver, providing better initialization to the gradient-based method and allowing explicit handling of state constraints. The sampling-based planner also allows direct integration of a particle filter, which is here used for online contact mode estimation. The approach is shown to improve performance in MuJoCo environments and the effects of problem stiffness and planing horizon are investigated. The estimator and planner are then applied to an impedance-controlled robot, showing a reduction in solve time in contact transitions to only gradient-based.    
\end{abstract}

\section{Introduction}\label{Introduction}
Contact-rich manipulation tasks such as door opening and gear mating involve significant environmental dynamics and constraints, which discretely switch during the task as contact conditions change. Manipulation methods which can maintain safety (e.g. limits on contact force and actuation) and robustness (variation in contact geometry or dynamics) are needed which are suited to discontinuous dynamics. 

Gradient-based optimization methods are popular for contact planning, with a range of formulations from contact implicit \cite{anitescu1997a, posa2014, lecleach2024, aydinoglu2022a} to hybrid contact modes \cite{jin2024, hogan2020a}. Sampling-based methods have also been used for higher-level contact mode planning \cite{chen2021, wu2020a} and continuous planning \cite{pang2023}.

The tradeoffs between sampling and gradient-based planners has led to recent interest in combining them, such as adding gradient steps to the sampling-based Cross-Entropy Method (CEM) \cite{bharadhwaj2020a, huang2021a}.  In contact, recent work has compared the variance of gradient-based and gradient-free policy learning in contact \cite{suh2022c}. This raises the question if a planning method can effectively use both sampling and gradients, where sampling may address the initialization sensitivity of gradient-based methods \cite{onol2020}, while gradient-based methods can provide explicit safety constraints. This paper proposes a planning method which leverages CEM and interior-point optimization for contact-rich planning.

\subsection{Related works}
Model-based planning for contact-rich robotic tasks finds the action sequence that minimizes a cumulative cost function, according to learned, parametrized, or simulated dynamics. Usually, optimized actions are executed with model predictive control (MPC), replanning at each time step. The problem can be formulated using complementary conditions for contact constraints \cite{stewart2000, anitescu1997a}, where sequential quadratic \cite{posa2014} or interior-point \cite{lecleach2024} solvers can be used. This contact-implicit trajectory optimization allows planning of contact mode and trajectory in one problem with multiple contacts and Coulomb constraints \cite{aydinoglu2022a, lecleach2024, onol2020}. Alternatively, hybrid dynamics can be written with dynamics for each contact mode \cite{jin2024, hogan2020a}. 

The contact planning problem is challenging, typically needing relaxations \cite{pang2023, lecleach2024}, linearization \cite{aydinoglu2022a, onol2020}, or quasi-static assumptions \cite{hogan2020a} to achieve acceptable solve times. The resulting problems can be sensitive to initialization \cite{onol2020}, and in general gradient-based methods can suffer from local minima or poor convergence on non-smooth problems, e.g. Newton's method convergence rate scales with the condition number of the Hessian \cite{rawlings2017}. 

\begin{figure}
    \includegraphics[width=\columnwidth]{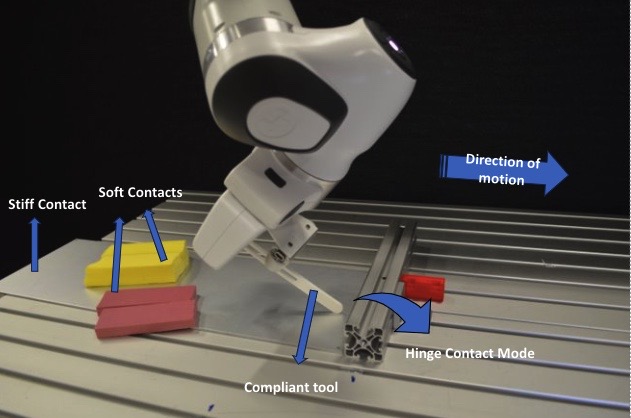}
    \caption{Experimental setup showing a robot making contact with a table, moving towards a hinge contact condition, providing a task with three modes: free space, vertical stiff contact, and hinge contact.  The soft contacts are used for validating the effect of contact stiffness on performance.\label{fig:exp_setup}}
\end{figure}

On the other hand, population-based sampling planners can parallelize well and do not need gradients, such as Random-Sampling Shooting \cite{Nagabandi2017}, Model Predictive Path Integral Control \cite{Williams2017}, and the Cross-Entropy Method (CEM) \cite{Rubinstein97}. Sampling-based planners have been used in contact planning, for higher-level mode initialization of a contact-implicit \cite{chen2021} and reachability planner \cite{wu2020a}. Sampling can also be used for the continuous planning, where Rapidly-exploring random trees have been applied to relaxed contact models \cite{pang2023}.

These zeroth-order optimizers maintain a sampling distribution action sequences \cite{okada2020}, which is sampled from then updated at each iteration, assigning a higher probability to higher reward action sequences. By adding temporally correlated actions and memory, CEM can be improved in both sample-efficiency and reward performance \cite{pinneri2021}. CEM can also be applied to interaction dynamics learned from an ensemble of Neural Networks \cite{Roveda2020ModelBasedRL} to realize online planning. 

However, CEM, like all sampling-based planners, suffers from scalability as the dimension of the action space or planning horizon grows, requiring more samples for convergence and having increasing reward variance \cite{bharadhwaj2020a, chua2018b}. Additionally, sampling with general state constraints requires either rejection sampling \cite{ichter2018} or smoothing \cite{pang2023}, raising challenges for efficient handling of contact constraints.

Recent work has combined zeroth-order and first-order planning methods. Authors in \cite{bharadhwaj2020a, huang2021a} proposed a combined planner that performs first-order gradient updates on the sampled trajectories. While this improves CEM performance, these methods still show scalability issues with high dimensional planning problems and don't handle constraints. For policy learning in contact, recent work \cite{suh2022c} has shown that first order methods can have higher variance than zeroth order methods at higher stiffness and in longer planning horizons. 
\begin{figure}
    \includegraphics[width=\columnwidth]{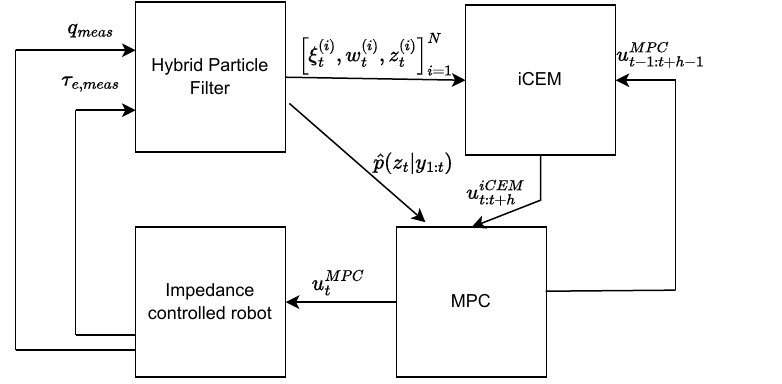}
    \caption{Block diagram of the proposed approach \label{fig:IntegrationDiag}}
\end{figure}
\subsection{Paper contribution}
To address poor convergence and local minima problems of gradient-based optimizers when dynamics are discontinuous, we propose a hybrid planner that uses both sampling- and gradient-based MPC. Contact is modelled with hybrid dynamics, where dynamics are switched according to contact mode. A particle filter is used to estimate the hybrid state online, providing a belief of current mode for the planner. The particle filter provide weights and particles to a CEM planner, which is used to initialize the action and state trajectory, modified to normalize sample cost by the weight and minimizing expected cost over the belief in mode. A gradient-based optimizer is then used, enforcing explicit constraints on the robot trajectory, as seen in Figure \ref{fig:IntegrationDiag}. 

Compared to the work of \cite{pinneri2021}, this CEM rolls out trajectories on a hybrid system, leveraging the contact mode belief provided by the filter to weight the computed samples costs. Compared to other combined sampling- and gradient-based planning approaches \cite{bharadhwaj2020a, huang2021a}, this  method can enforce explicit constraints on the robot trajectory and handles hybrid dynamics. Our method is also validated experimentally, showing the feasibility of the approach with a real contact task as seen in Figure \ref{fig:exp_setup}. The ability to estimate contact mode online, as well as safely respond with changes in control is validated. 

This paper first introduces the dynamic models used, estimation of contact mode. Then, planning framework is introduced and studied in simulation. Finally, experiments are used to validate the approach, comparing the proposed planner to only MPC and only CEM. We find that the CEM initialization helps with discontinuous dynamics, reducing the mean and variance of the total solve time, even from just one iteration, as compared to a pure gradient-based MPC for contact. 

\section{Contact, Dynamics and Observation Models}
This section introduces the dynamic models used for contact and the hybrid dynamic model. 

\subsection{Stiffness Contact Model}
We model environment contact as parallel 1-DOF stiffnesses to compromise between model identification efficiency, ease of deployment, generalizability, and differentiability \cite{castro2022a, haninger2023a}. The $i^{\mathrm{th}}$ stiffness element exerts contact force $F_i$ at the point of contact of
\begin{equation}
    F_i = K_i^T\left(x_i^o-(R x_i+x)\right),  \label{env_contact}
\end{equation}
where stiffness $K_i\in\mathbb{R}^3$ has a rest pose at $x_i^o\in\mathbb{R}^3$ and $x_i\in\mathbb{R}^3$ is the contacting point in tool center point (TCP) frame, transformed into world coordinates with $R$ and $x$ , the orientation and position of the end-effector. The contact normal and stiffness are jointly described by the vector $K_i$, \textit{i.e.}, the contact normal is $n_i = K_i/\Vert K_i \Vert$, and the stiffness is $\Vert K_i\Vert$. For a single point stiffness, the joint torque induced is $\tau_i = J_i^TF_i$, where $J_i = \partial (R(q)x_i+x(q)) / \partial q$.

When $N_c$ contacts are active, the total external joint torque is $\tau_e = \sum_{i=1}^{N_c} \tau_i(q)$. The parameters $K_i$, $x_i^o$, and $x_i$ can be identified from least-squares fit on a dataset as  $\min_{K,x^o, x} \sum_t \Vert \tau_t^m - \tau_e(q_t, K, x^o, x)\Vert$.  These contact forces are composed of differentiable operations, and easily implemented in automatic differentiation (AD) frameworks to support their use in gradient-based planning.

\subsection{Discontinuous Dynamics}
The standard serial manipulator robot dynamics 
\begin{equation}
    M(q)\ddot{q}+ C(q,\dot{q}) + B\dot{q} + G(q) = \tau_m + \tau_e \label{inertial_dyn}
\end{equation}
is assumed to be available in an AD framework, with joint position $q\in\mathbb{R}^n$, inertia matrix $M(q)$, Coriolis terms $C(q, \dot{q})$, viscous damping $B$, gravitational torque $G(q)$, input torque $\tau_m$, and torques from contact $\tau_e$. 

We discriminate between different contact modes by letting the torque $\tau_e$  depend on the mode $z\in [1,\dots, N_z]$, where each mode has different point contact models, resulting in a different $\tau_e$. We suppose the mode changes during the task, denoting the value at $t$ as $z_t$. As typical in contact models, we use semi-implicit integration which handles stiff differential equations better \cite{stewart2000}. For a time step $\Delta_t$
\begin{align}
    z_t & \sim p(z_t|z_{t-1})\\
         \xi_t = \begin{bmatrix} q_t \\ \dot{q}_t \end{bmatrix} & = \begin{bmatrix} q_{t-1} +\Delta_t\dot{q}_t \\ \dot{q}_{t-1} +\Delta_tM^{-1}(\tau_m+\tau_e(z_t)-B\dot{q}_t) \end{bmatrix}
\end{align}
where $\xi$ are the joint states and $p(z_t|z_{t-1})$ is a diagonal-dominant discrete transition matrix. Here, the transition matrix is assumed to be a fixed prior, tuned empirically, but could reflect a prior belief about mode transitions based on position or camera data.

As the external contact force $\tau_e(z_t)$ depends on the current discrete mode $z_t$, the dynamics can be written as
\begin{equation}
     \xi_t = \begin{bmatrix} q_{t} \\ \dot{q}_{t}\end{bmatrix} = f\left(\begin{bmatrix} q_{t-1} \\ \dot{q}_{t-1}\end{bmatrix}, z_{t-1}, \tau_{m,t-1} \right). \label{eq:nonlin_dyn}
\end{equation}

For an extended Kalman filter (EKF), the dynamics in \eqref{eq:nonlin_dyn} are linearized with respect to $\xi$, adding a process noise contribution of $w\sim\mathcal{N}(0, Q)$. The observation model would comprise the external joint torques $\tau_e$ plus the joint positions as
\begin{align}
    y^z & = C^z\xi + v \label{eq:obs_model},
\end{align}
where $C^z=[C_q^T, \frac{\partial \tau_e(z)}{\partial \xi}^T]^T$, $C_q = [I_n, 0_n]$, $v\sim\mathcal{N}(0,R)$ is the measurement noise, assumed to be independent and identically distributed, $\partial \tau_e(z) / \partial \xi$ is the Jacobian of the external contact torques $\tau_e(z)$ with respect to the joint states $\xi$, and $I_n$ an identity matrix of dimension $n$.

\subsection{Robot impedance dynamics}
The robot is controlled with a Cartesian space impedance controller, where the $3$ DoF impedance controlled robot dynamics are
\begin{equation}
   \tau_m = J^T(q)\left[D\dot{x}+K\left(x-x_0\right)\right] + G(q) + C(q, \dot{q})
   \label{imp_dyn}
\end{equation}
where $x, \, \dot{x} \in \mathbb{R}^3$ are respectively the TCP Cartesian position and velocity, $J=\partial x / \partial{q}$ and the impedance parameters are $x_{0} \in \mathbb{R}^3$ which represents the Cartesian impedance rest position, the damping $D \in \mathbb{R}^{3\times3}$, and stiffness $K \in \mathbb{R}^{3\times3}$ matrices, which are here all diagonal.

The dynamics in Eq.\ref{imp_dyn} can then be updated to provide controlled discretized dynamics of
\begin{equation}
    \begin{bmatrix} q_{t} \\ \dot{q}_{t}\end{bmatrix} = f\left(\begin{bmatrix} q_{t-1} \\ \dot{q}_{t-1}\end{bmatrix}, x_{0,t-1}, z_{t-1} \right). \label{eq:nonlin_mpc_dyn}
\end{equation}

\section{Online Planning Method}
This section introduces the belief estimation, cross-entropy and gradient-based planning method used.
\subsection{Belief estimation}
For estimating the hybrid state, we adopt a hybrid particle filter algorithm \cite{thrun2002}. The proposed algorithm maintains an overall belief of the discrete mode $\hat{p}(z_t|y_{1:t})$, from the weighted belief of individual particles $\hat{p}(z^{(i)}|y_{1:t-1})$. In each step, each particle samples a discrete mode $\hat{z}_t^{(i)}\sim \hat{p}(z_t^{(i)}|y_{1:t-1})$, then propagates the mean $\mu^{(i)}_t$ and the covariance $\Sigma^{(i)}_t$ of continuous state $\xi_t$ with the corresponding mode dynamics and observation equations via a standard Extended Kalman filter step \cite{thrun2002}. Each particle $p^{(i)}$  is characterized by the tuple $[\mu_t^{(i)}, \Sigma_t^{(i)}, \hat{p}(z^{(i)}_t|y_{1:t-1}), \hat{z}_t^{(i)}]$.
An overview can be seen in Algorithm \ref{alg:hyb}, where $\hat{z}^{(i)}_t$ is the sampled mode for particle $i$, $\hat{S}^{(i)}_t = \mathrm{cov}(y_t|y_{1:t-1})$ the predicted observation covariance, $w^{(i)}_t$ the weight, $\hat{y}^{(i)}_{t|t-1}=\mathbb{E}(y_t|y_{1:t-1})$ is the predicted measurement, and $\delta$ is the Dirac delta function.
\begin{algorithm}
\caption{Hybrid particle filter, changes to PF \textcolor{blue}{in blue}}\label{alg:hyb}
\begin{algorithmic}
\ForAll{particles $p^{(i)}$}
\State $ \hat{\mu}^{(i)}_{1|0}\leftarrow  \mu_0$; \,\, $ \hat{\Sigma}^{(i)}_{1|0}\leftarrow  \Sigma_0$; \,\, $ \hat{p}(z^{(i)}_{1}|y_{1})\leftarrow p(z_0)$
\EndFor
\ForAll{time step $t$}
\ForAll{particles $p^{(i)}$}
{\setstretch{1.4}
\State $\hat{p}(z_t^{(i)}|y_{1:t-1})=\hat{p}(z^{(i)}_{t-1}|y_{1:t-1})p(z_t|z_{t-1})$
\State \textcolor{blue}{$\hat{z_t}^{(i)}\sim \hat{p}(z_t^{(i)}|y_{1:t-1})$}
\State \textcolor{blue}{EKF update $(\hat{y}^{(i)}_{t|t\mathtt{-}1},\hat{S}^{(i)}_t,\hat{\mu}^{(i)}_t,\hat{\Sigma}^{(i)}_t)$} %
\State $w^{(i)}_t\propto \mathcal{N}(y_t;\hat{y}^{(i)}_{t|t-1},\hat{S}^{(i)}_t)$
\EndFor
\EndFor
\State Resample particles $\{p^{(i)}, w_t^{(i)}\}_{i=1}^N$
\State \textcolor{blue}{Estimate belief ${\hat{p}(z_{t}|y_{1:t}) \propto \sum_{i=1}^N w^{(i)}_{t}\delta_{z_{t}}(\hat{z}^{(i)}_{t})}$}
}
\EndFor
\end{algorithmic}
\end{algorithm}
The posterior belief $\hat{p}(z_t|y_{1:t})$ is used to detect changes in contact, and is used in the planning methods to weight the costs associated with different modes.

\subsection{Proposed Cross-Entropy Method}
The Cross-Entropy Method (CEM) samples actions from an action distribution,  rolls out these trajectories and evaluates their resulting costs, adapting the action distribution to reduce loss. Sampling-based approaches may avoid local minima problems or convergence issues when gradients are not smooth due to the discontinuous dynamics, potentially reaching state-action pairs that would not be reached by gradient-based methods. Thus, we use CEM iterations before proceeding with a gradient-based non-linear solver.  

The adopted CEM algorithm is based on the work of \cite{pinneri2021}.  The CEM here is initialized from the hybrid filter at each time step $t$ with the set of particles consisting of continuous joint states values $\xi_t^{(i)}$,  their weight $w_t^{(i)}$ and  discrete sampled state $z^{(i)}_t$,
\begin{equation}
     \left[\xi^{(i)}_t,w_t^{(i)},z^{(i)}_t\right]_{i=1}^N.
\end{equation}
This list is used for calling the dynamics with the hybrid mode $z_t^{(i)}$ with initial continuous state $\xi_t^{(i)}$ along with the sampled actions $u_{t:t+h}$.  As in typical CEM, actions are iteratively sampled and the sampling distribution shifted towards the elite samples which have the best loss.  A detailed explanation of the steps is given in Algorithm \ref{alg:prop_iCEM}. 
\begin{algorithm}
\caption{Proposed iCEM, changes to \cite{pinneri2021} \textcolor{blue}{in blue}}
\label{alg:prop_iCEM}
\begin{algorithmic}
\Require {$N$: number\ of \ particles,\ $h$: planning\ horizon,\ 
$d$: action\ dimension,\ $\beta$: colored\ noise\ exponent,\ $N_{iter}$: CEM-iterations, $\left[\xi^{(i)}_t,\textcolor{blue}{w_t^{(i)},z^{(i)}_t}\right]_{i=1}^N$: list\ of\ particles,\ $K$: elite\ size} \\
\hspace*{-1.25em} \textbf{Initialize:}
\State $u_{t:t+h} = u_{t-1:t+h-1}^{MPC}$
\State $\Sigma^u = \textbf{1}$
\\ \hspace*{-1.25em} \textbf{Loop:}
{\setstretch{1.4}
\For{$k$\ in\ $N_{iter}$}
\State $u^{(i)}_{t:t+h} \leftarrow \mathtt{clip}(u_t + C^\beta(d, h)\cdot \Sigma^u)$
\State $ \xi^{(i)}_{t+1:t+1+h} \leftarrow f(\xi^{(i)}_t,u^{(i)}_{t:t+h},\textcolor{blue}{z^{(i)}_t})$
\State ${l}^{(i)} \leftarrow \sum_{\tau=t+1}^{t+1+h} l(\xi^{(i)}_{\tau}, u^{(i)}_{\tau},\textcolor{blue}{z^{(i)}_t} )$
    \State \textcolor{blue}{${l}^{(i)} \leftarrow \frac{{l}^{(i)}}{w^{(i)}_t}$}
\State $ u_{t:t+h}, \Sigma^u \leftarrow $ \ fit\ Gaussian\ distribution\ to\ elite set
\EndFor \\
}
\hspace*{-1em}\textbf{return} best action trajectory $u_{t:t+h}^{iCEM}$
\end{algorithmic}
\end{algorithm}
In Algorithm \ref{alg:prop_iCEM}, $u_{t:t+h}$ and $\Sigma^u$ represent the mean and the standard deviation of the action sequences, at the beginning of each time step $t$ we initialise the mean with the optimal value found at the previous time step $t-1$ by the non linear MPC solver while the standard deviation of the sequences is re-initialised every time with the matrix $\textbf{1} \in \mathbb{R}^{d\times h}$, whose elements are ones. $C^\beta(d, h)$ represents the colored noise distribution, built following the approach in \cite{pinneri2021}. Within the CEM inner loop, we keep updating the action sequences statistics by fitting at each iteration the best $K$ trajectories; as long as the solver converges the mean $u_{t:t+h}$ is initialized at each time step with the optimal solution and the likelihood of sampling optimal solutions increases. The cost of each particle $l^{(i)}$ is divided by the particle weight, so a low likelihood particle results in a higher cost, to avoid actions which favor low-likelihood particles from being sampled more heavily. The cost function itself is the same as described in the following section for the MPC problem.

The CEM returns both the best action trajectory, along with the associated state trajectory, which are used to initialize the decision variables in a gradient-based MPC problem. 

\subsection{MPC problem}
The MPC problem is formulated using multiple-shooting transcription with a generic problem statement of
\begin{align}
 \min_{u_{t:t+h}} & \sum_{t}^{t+h}\sum_{z=1}^{N_z} \hat{p}(z|y_{1:t})l(\xi^{z}_t,u_t, z) & \label{eq:mpc_prob} \\
 \mathrm{s.t.}\,\,\,& \forall z \in [1,\dots,N_z],\quad \tau \in [t,\dots,t+h-1]: \\
 &\Vert \xi^{z}_{\tau+1}-f(\xi^{z}_{\tau},u_{\tau}, z)\Vert \leq \rho \\
 &g(\xi_t^z,u_t) \geq 0 \\
 &u \in \mathcal{U}
\end{align}
where $h$ is the planning horizon, $\mathcal{U}$ is the range of allowed inputs, $\rho$ is the slack for continuity constraints (the inequality is applied element-wise), $z \in [1,\dots,N_z]$ represents the set of the different contact modes, and $g$ represent general state and input dependent inequality constraints. A trajectory is rolled out for each mode, $\xi^z_t$, where continuity is imposed according to that mode's dynamics in $f$; $\hat{p}(z|y_{1:t})$ is the contact state belief information coming from the hybrid filter at time step $t$. The constraints shown in the optimization problem are nonlinear, so an interior-point nonlinear optimization solver is used. 

The general stage cost function, which would be used also in the CEM loop, is defined as
\begin{equation}
\begin{split}
    l(\xi^{z},u, z) =  & (x_{d,z}-x)^TQ_x(x_{d,z}-x) \\ &+(u-x)^TQ_u(u-x)+\dot{x}^TQ_{\dot{x}}\dot{x}
\end{split}
\end{equation}
where $x_{d,z}$ is a desired position in world frame to be tracked for the $z$ mode, $x$ and $\dot{x}$ are the Cartesian position and velocity of the robot TCP expressed with the forward kinematics function, directly depending on the joint configuration. In this scenario, the state $\xi^{z}=(q^{z},\dot{q}^{z})$ is represented by the robot joint variables for each mode $z$, while the adopted control input $u$ is shared among the different modes and is impedance rest position $x_0$. The constraint is $g = \overline{F}_{imp}-\Vert K(x - x_0) \Vert_2$, limiting the virtual impedance force applied to $\overline{F}_{imp}$.

\section{Simulation Studies}
To study the proposed approach, studies in simulation are used to investigate the performance in terms of cost and solve time.

\subsection{MuJoCo Gym}
To compare the methods on problems with contact, they are compared on the MuJoCo Gym environments in Brax \cite{freeman2021}. No mode estimation with a particle filter is used; the solver cost is directly the negative reward, no constraints are applied, a maximum episode length of $500$ is used, and the episode can terminate early according to the environment.  A planning horizon of $20$ is used, with CEM parameters of $N_{iter}=150$, $N=128$, $K=10$, and $10$ random seeds used for each solver.

\begin{figure}
    \includegraphics[width=\columnwidth]{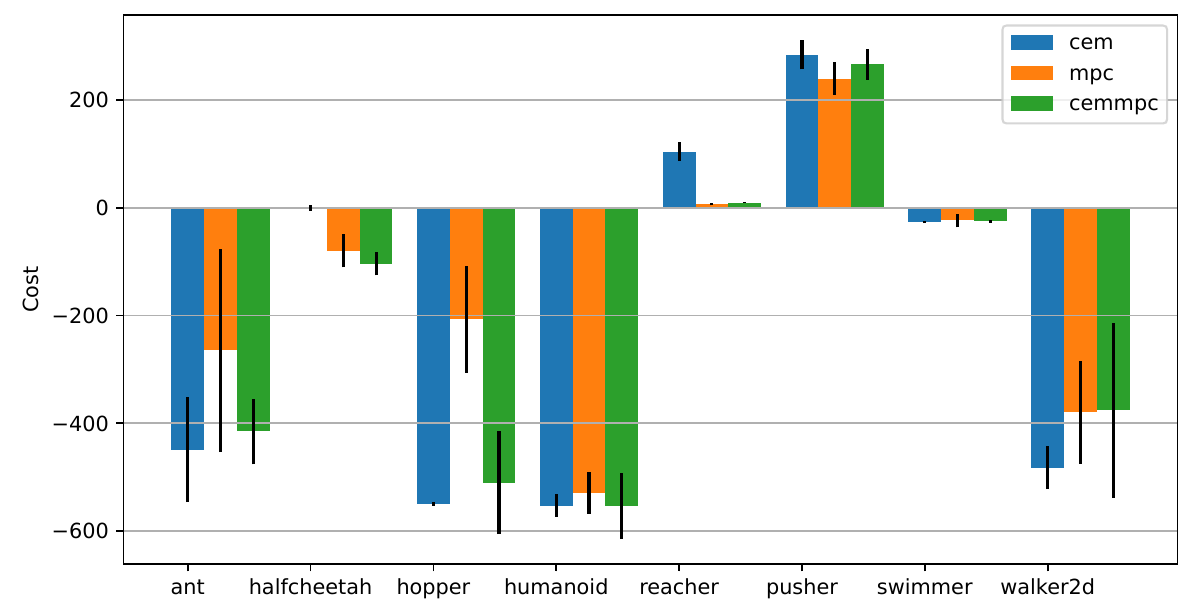}
    \caption{Performance on the MuJoCo environments with planning horizon $h=20$.  .\label{fig:solver_mujoco_comp}}
\end{figure}

The results can be seen in the attached video as well as Figure \ref{fig:solver_mujoco_comp}.  In `ant' and `hopper', CEM performs better than MPC, and the proposed approach matches CEM performance. In `halfcheetah', the proposed approach improves over both MPC and CEM,  and in other environments, the performance of the proposed approach is similar to MPC. As the MPC solver terminates successfully in these problems (i.e. relative tolerance of $1e-8$ is met), this suggests that a CEM initialization does not hurt performance, and can in some problems lead the MPC to a better local minima.

\subsection{CEM vs MPC in stiff contact}
\begin{figure}
    \includegraphics[width=\columnwidth]{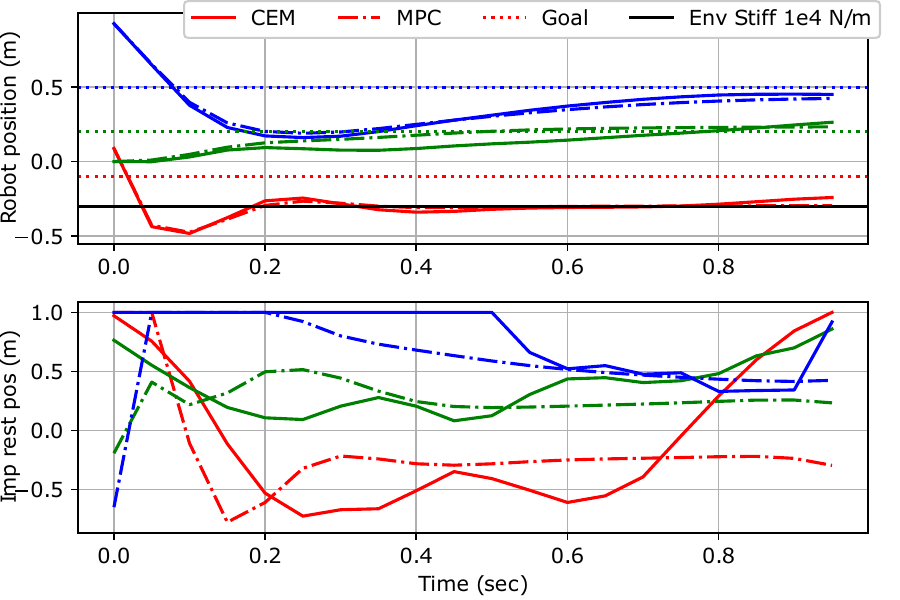}
    \caption{Comparing state and control trajectory from CEM vs MPC for stiff dynamics, the x-, y-, and z-positions shown in red, green, blue, respectively.\label{fig:mpc_vs_cem_stiff}}
\end{figure}

To contrast with robot manipulator dynamics, a cartesian position tracking problem is set for the a simulated impedance-controlled Franka Emika robot in contact with stiffness $K=1e4$ N/m.  To see the variance in the performance, each problem is solved repeatedly with the initial state $\xi_t$ perturbed with $\mathcal{N}(0,0.1)$ between solves, finding the mean and covariance. As can be seen in Figure \ref{fig:mpc_vs_cem_stiff}, the CEM (solid line) has more variation in the control trajectory, and the state trajectory is slower to converge to the final position the MPC (dash-dotted line). 

To study the relative performance of the gradient- and sampling-based methods in discontinuous dynamics, we examine two performance characteristics: time to solve and resulting expected cost. First, the solver is used to solve a free-space problem according to \eqref{eq:mpc_prob} with horizon $h=20$ steps and stepsize $\Delta_t=0.05$ seconds. Then, the dynamics is changed to a single point contact model of stiffness $K=1e4$ N/m and the previous solution used initialize the current step. This initialization is then used for $N_{iter}$ iterations of the CEM before starting the next MPC solve.  

\begin{figure}
\centering
    \includegraphics[width=0.95\columnwidth]{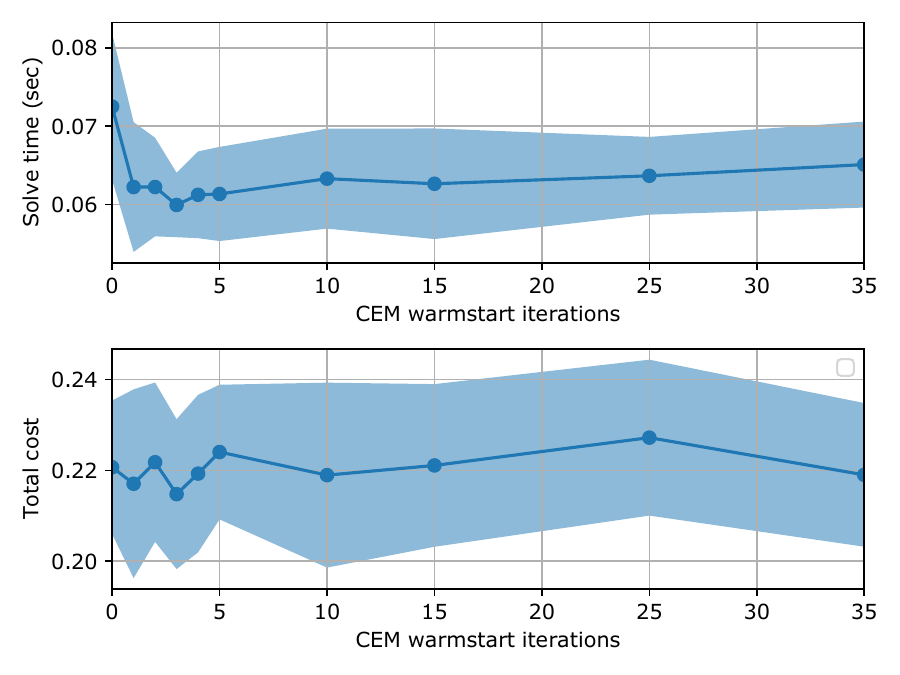}
    \caption{Effect of CEM warmstart iterations on total solve time and trajectory cost for proposed approach on trajectory tracking problem.  The CEM does not significantly change performance, but can improve solve time. \label{fig:cem_warmstart_num_iter}}
\end{figure}

The mean and covariance of the total solve time for the CEM iterations and MPC solve can be seen in Figure \ref{fig:cem_warmstart_num_iter}. As can be seen, the CEM helps to reduce both the mean and variance of the solve time, even from just one iteration.  However, as the number of CEM iterations increases, the time required increases, indicating the additional time required to do more CEM iterations is not saving as much time in the MPC solver. On this problem, the number of CEM iterations do not significantly affect the total cost.

\begin{figure}
\centering
    \subfloat[Effect of stiffness]{\includegraphics[width=0.97\columnwidth]{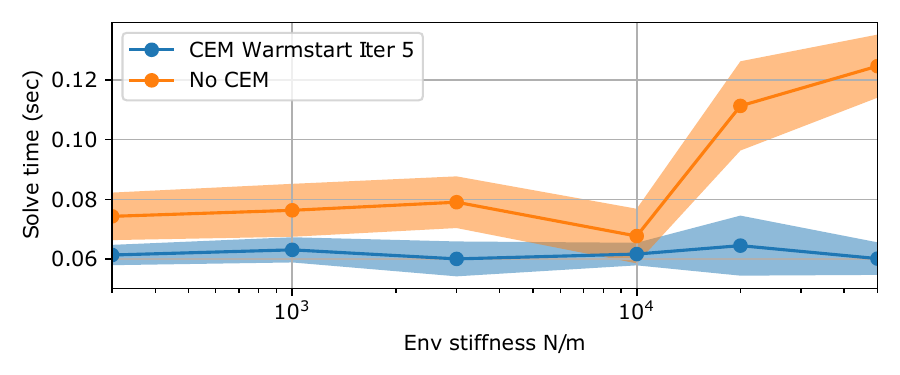}} \\
    \subfloat[Effect of planning horizon]{\includegraphics[width=0.97\columnwidth]{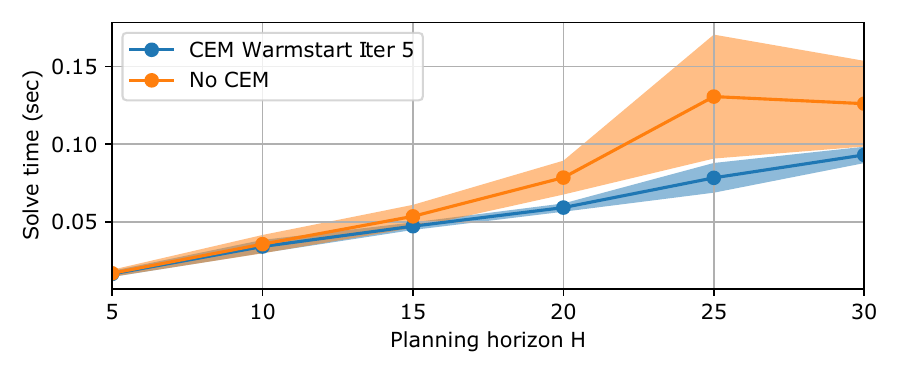}}
    \caption{Effect of problem parameters (stiffness and planning horizon) on solve time for only MPC (orange) and proposed approach (blue).  \label{fig:cem_warmstart_stiff_and_H}}
\end{figure}

To analyze the impact of important process parameters, we study the improvement by $N_{iter}=5$ warmstarts over a range of stiffnesses and planning horizons. The results can be seen in Figure \ref{fig:cem_warmstart_stiff_and_H}, where as the stiffness and horizon increase, the CEM improves both the mean and variance of solve time.  This trend matches the impact of stiffness and planning horizon in policy gradient estimation \cite{suh2022c}, where gradient-free methods performed better at long horizons and higher stiffnesses. 

\section{Experiments}
To validate the performance on real-world contact tasks, we apply the proposed method to the two tasks seen in Figure \ref{fig:exp_setup}.  The code and data are available at \url{https://gitlab.cc-asp.fraunhofer.de/hanikevi/contact_mpc}.  

\subsection{Vertical contact task}
\begin{figure}
    \includegraphics[width=\columnwidth]{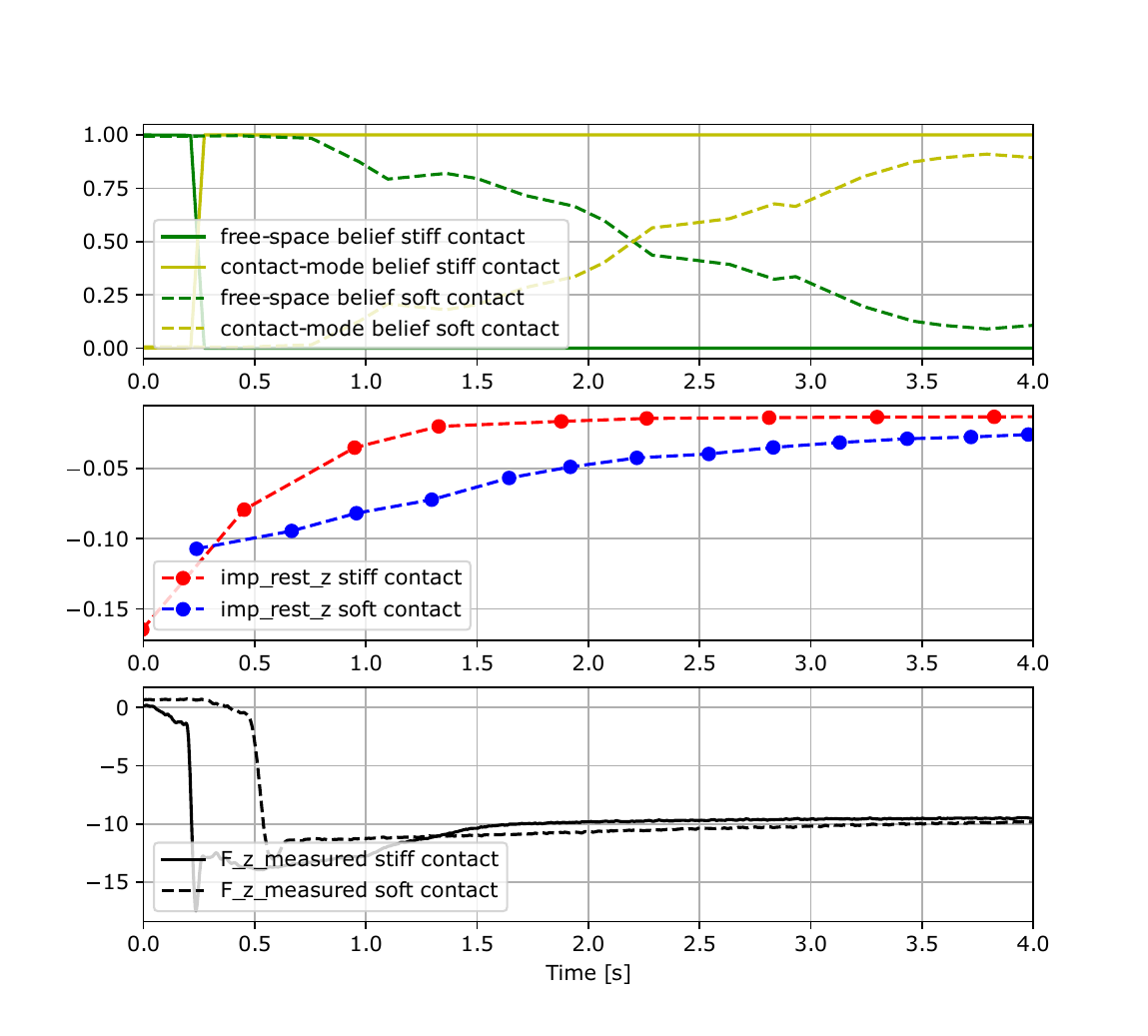} 
    \caption{Effect of stiffness on contact detection \label{fig:exp_vertical}}
\end{figure}

In the vertical contact task, the robot moves from free space into contact with an environment. The stiffness of the environment is adjusted by adding the yellow or purple foam seen in Figure \ref{fig:exp_setup}, or direct contact with the aluminum table. In Figure \ref{fig:exp_vertical} we can see how the stiffness parameter influences the contact mode detection responsiveness. Despite the developed CEM+MPC approach successfully worked for all the contact environments, from the plot we can see that with stiffer contacts the mode detection is more sharp, with the transition triggered right in correspondence of the vertical force peak. For lower stiffness contacts, mode detection is slightly delayed with respect to the force peak, probably because the lower measured interaction force makes it more difficult for the filter to assign different likelihood values.
\subsection{Pivot task}
\begin{figure}
    \subfloat[Simple gradient-based MPC]{\includegraphics[width=\columnwidth]{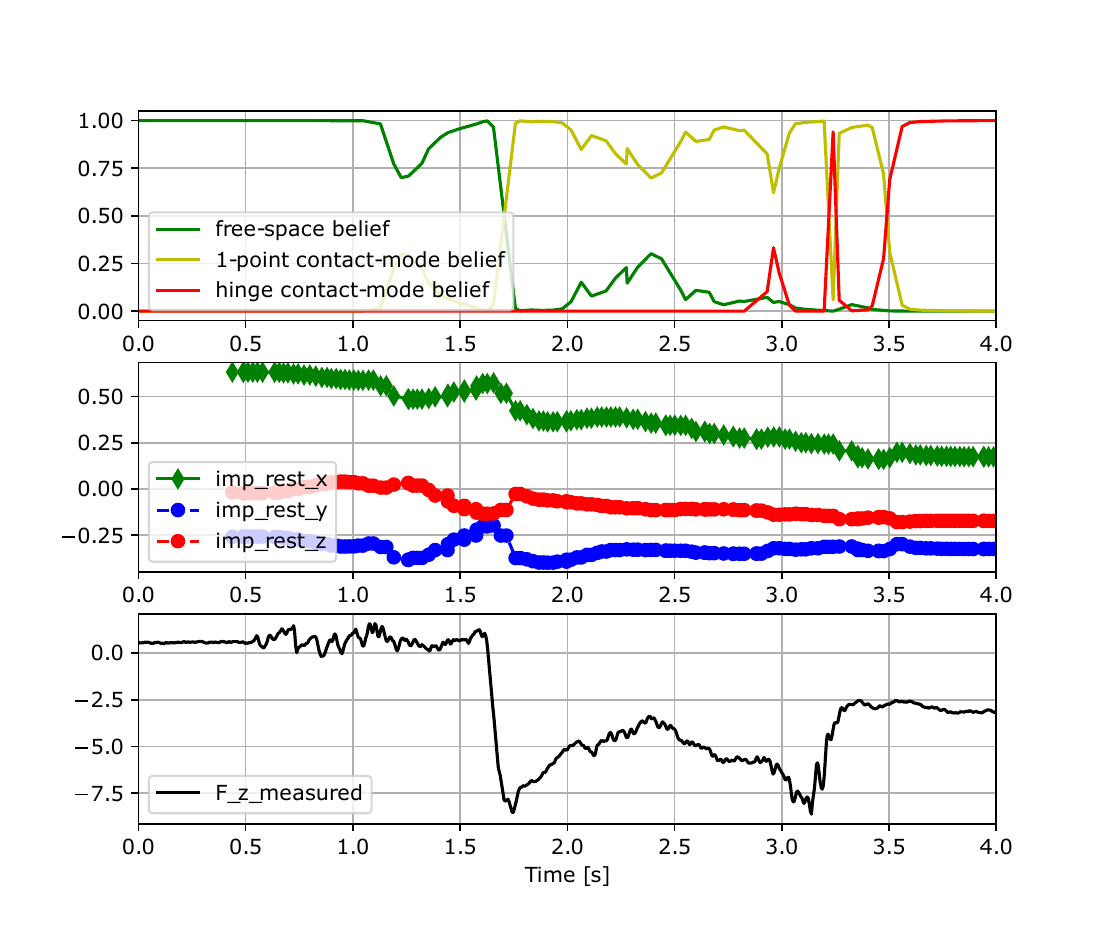}} \\
    \subfloat[Proposed CEM+MPC approach]{\includegraphics[width=\columnwidth]{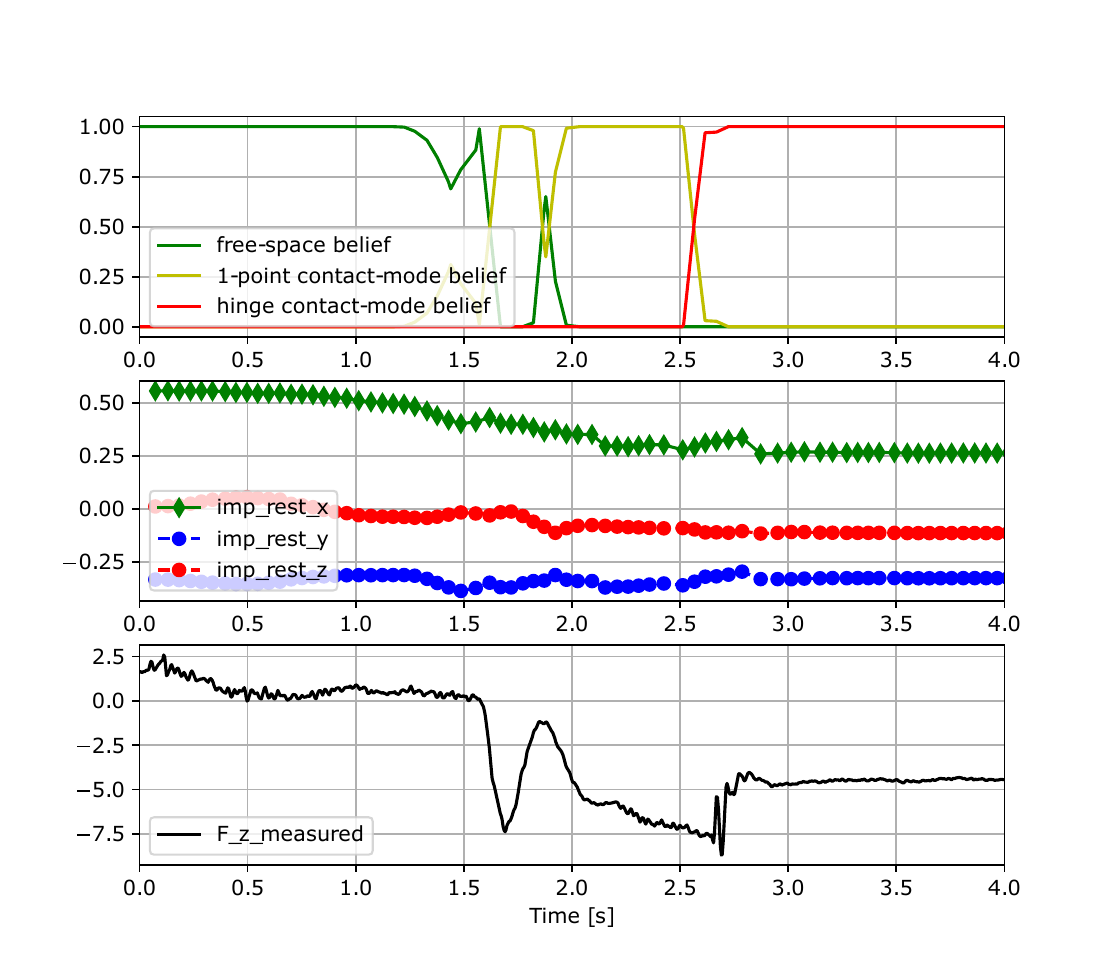}}
    \caption{Comparison between the simple MPC and CEM+MPC approaches for the pivot task.\label{fig:exp_pivot}}
\end{figure}

In this task, the robot makes contact with the table, slides along the surface until making contact with the wall. The MPC problem is solved with $h=13$, $\Delta_t=0.04$, an impedance cost of $Q_u=0.05$, maximum impedance force of $\overline{F}_{imp}= 25$, a desired pose of $x^d=[-,-,0.01]$ in free space, and $x^d=[0.35, -, 0.01]$ in plane and hinge modes, where $-$ are omitted from the cost. In this experimental setup, the proposed approach was compared to a simple MPC approach without the CEM warm-start, in order to validate experimentally the results achieved in simulation. From the plots in Figure \ref{fig:exp_pivot}, it can be seen that without the CEM warm-start, the impedance rest position shows a small discontinuity in its trend, and this can be noticed especially when the first transition is triggered. This result is in line with what we have shown in the simulation study, the proposed CEM+MPC approach helps in decreasing the total solve time at contact transitions.

\section{Conclusion}
This paper proposed an approach to planning on discontinuous dynamics which uses a modified CEM to initialize a gradient-based planner. It was shown that such an approach can reduce the variation in solve time when discontinuities in dynamics occur. The approach was also validated on real contact tasks, showing that planning method results in safe robot trajectories which respond to the contact conditions.
\bibliographystyle{IEEEtran}
\bibliography{lib_kevin, biblio}

\end{document}